\documentclass{article}

\PassOptionsToPackage{numbers, compress}{natbib}

\usepackage[utf8]{inputenc} 
\usepackage[T1]{fontenc}    
\usepackage{hyperref}       
\usepackage{url}            
\usepackage{booktabs}       
\usepackage{amsfonts}       
\usepackage{nicefrac}       
\usepackage{microtype}      
\usepackage{bm}
\usepackage{subcaption}
\usepackage[table]{xcolor}
\usepackage[most]{tcolorbox}
\usepackage{amsthm}
\usepackage{multirow}
\usepackage{siunitx}
\usepackage{tikz}
\usepackage{hhline}
\usetikzlibrary{3d,decorations.text,shapes.arrows,positioning,fit,backgrounds}
\tikzset{pics/fake box/.style args={
#1 with dimensions #2 and #3 and #4}{
code={
\draw[gray,ultra thin,fill=#1]  (0,0,0) coordinate(-front-bottom-left) to
++ (0,#3,0) coordinate(-front-top-right) --++
(#2,0,0) coordinate(-front-top-right) --++ (0,-#3,0) 
coordinate(-front-bottom-right) -- cycle;
\draw[gray,ultra thin,fill=#1] (0,#3,0)  --++ 
 (0,0,#4) coordinate(-back-top-left) --++ (#2,0,0) 
 coordinate(-back-top-right) --++ (0,0,-#4)  -- cycle;
\draw[gray,ultra thin,fill=#1!80!black] (#2,0,0) --++ (0,0,#4) coordinate(-back-bottom-right)
--++ (0,#3,0) --++ (0,0,-#4) -- cycle;
\path[gray,decorate,decoration={text effects along path,text={CONV}}] (#2/2,{2+(#3-2)/2},0) -- (#2/2,0,0);
}
}}
\tikzset{circle dotted/.style={dash pattern=on .05mm off 2mm,
                                         line cap=round}}

\theoremstyle{definition}
\newtheorem{definition}{Definition}[section]

\theoremstyle{definition}
\newtheorem{goal}{Goal}[]

\usepackage{array}
\newcolumntype{P}[1]{>{\centering\arraybackslash}p{#1}}

\newcommand*{\eg}{\emph{e.g.} }
\newcommand*{\ie}{\emph{i.e.} }
\newcommand*{\etc}{%
    \@ifnextchar{.}%
        {etc}%
        {etc.}%
}

\definecolor{ao}{rgb}{0.0, 0.5, 0.0}

\newcommand*{\wrt}{w.r.t. }

\newcommand{\coolfont}[1]{{\fontfamily{cmtt}\selectfont #1}}
\usepackage[preprint]{neurips_2019}

\title{MonoNet: Towards Interpretable Models by Learning Monotonic Features}

\author{%
  An-phi Nguyen
  \\
  IBM Research Z\"urich, ETH Z\"urich\\
  \texttt{uye@zurich.ibm.com} \\
  \And
  Mar\'ia Rodr\'iguez Mart\'inez \\
  IBM Research Z\"urich \\
  \texttt{mrm@zurich.ibm.com} \\
\\
}

\begin{document}

\maketitle

\begin{abstract}

Being able to interpret, or explain, the predictions made by a machine learning model is of fundamental importance. This is especially true when there is interest in deploying data-driven models to make high-stakes decisions, e.g. in healthcare. While recent years have seen an increasing interest in interpretable machine learning research, this field is currently lacking an agreed-upon definition of interpretability, and some researchers have called for a more active conversation towards a rigorous approach to interpretability. Joining this conversation, we claim in this paper that the difficulty of interpreting a complex model stems from the existing interactions among features. We argue that by enforcing \emph{monotonicity} between features and outputs, we are able to reason about the effect of a single feature on an output independently from other features, and consequently better understand the model. We show how to structurally introduce this constraint in deep learning models by adding new simple layers. We validate our model on benchmark datasets, and compare our results with previously proposed interpretable models. 
\end{abstract}

\section{Introduction}

State-of-the-art deep learning networks are achieving strong predictive power,  but the gain in accuracy often comes at the price of transparency, and the decision reached lacks \emph{interpretability}.
Being able to interpret, or explain, the predictions made by a machine learning model is of fundamental importance, especially in sensitive domains, such as healthcare, crime recidivism or finance. If users do not trust a model, they will not use it, or even worse, they will use it and be inadvertently exposed to hidden biases~\cite{lipton2016mythos,wexler_opinion_2018}.
On the other hand, if the system can explain its reasoning, then  the soundness of the  reasoning  can be examined~\cite{doshi2017towards}.

It then comes as no surprise that an increasing number of researchers are focusing on either creating \emph{accurate} models that are already interpretable (\emph{ante-hoc} interpretability) \cite{angelino2017learning, yang2017scalable, melis2018towards} or models that can a posteriori provide explanations for already-trained black-box models (\emph{post-hoc} interpretability) \cite{ribeiro2018anchors,lundberg2017unified,ancona2017better,sundararajan2017integrated,shrikumar2017deeplift,ribeiro2016lime,selvaraju_grad-cam:_2016,bach_pixel-wise_2015,simonyan_deep_2013}.
However, requiring the model itself to be completely transparent can be too restrictive and may result in a model becoming too complex to be understood. 
On the other hand, a posteriori methods often derive local explanations, e.g. valid only around a particular example, due to the lack of access to the inner workings of the model. Furthermore, these methods often suffer  from problems related to the  definition of locality~\cite{alvarez-melis_causal_2017}, identifiability~\cite{li_deep_2017},  computational cost~\cite{yosinski_understanding_2015} and instability~\cite{alvarez-melis_robustness_2018} (see Section~\ref{sec:related}).

While a number of authors have made important contributions to this field, the community has still not agreed on  a \emph{definition of interpretability} \cite{doshi2017towards, gilpin2018explaining}, with consequent lack of standards to  evaluate different methods. These shortcomings are difficult to tackle and have prompted many researchers to ask for a discussion between all the \emph{stakeholders} \cite{lipton2017doctor} and including perspectives from different fields \cite{miller2018explanation}. 
Joining this discussion, this work aims to contribute towards finding a definition of interpretability that many can agree on. 
%
In this paper we propose a set of \emph{necessary} conditions that an interpretable model should abide to. 
One of these conditions is what \citet{lipton2016mythos} refers to as \emph{algorithmic transparency}, the opposite of opacity or blackbox-ness; it implies some level of understanding the mechanism by which the model works. \emph{(Human) Simulatability} is another important property of interpretable models, i.e.  the ability of a person to \emph{simulate} a model and get the correct output for a given input. The notion of simulatability is inspired by   sparse linear models, as produced by lasso regression~\cite{tibshirani_regression_1996}, which are more interpretable than dense linear models learned on the same inputs.


\section{Related work}\label{sec:related}

As mentioned previously, recently there has been an increasing volume of work in interpretable machine learning research. To understand the main advantages and drawbacks of the existing algorithms, we can categorize them in two classes: \emph{global} and \emph{local} methods. 

\paragraph{Global models} Global models are models that are fully \emph{transparent} in the sense that they provide the user with an overview of the whole decision process in terms of (possibly high-level) features, model weights and model parameters. Ideally, by having access to this information, a human should be able to completely simulate the decision process of the global model (\emph{simulatability}). Some examples of global models are decision trees \cite{quinlan1986decision}, rule lists \cite{letham2015interpretable} or risk score models \cite{ustun2016learning}. These methods suffer from two main drawbacks. Firstly, faced with a difficult learning problem, accurate models may become too complex to understand (\eg decision trees that become too deep). Secondly, some algorithms may suffer from a stability problem. For example, it is well known that decision trees are difficult to train since small perturbations in the training data may lead to different trees \cite{breiman1996bagging,li2002instable}.  

\paragraph{Local models} Local methods, on the other hand, aim to provide explanations that are valid only for the single sample at hand. From a human user perspective, local explanations are arguably more easily understood, since they usually involve only a few features at once. Examples of these models are \emph{backpropagation-like} methods \cite{sundararajan2017integrated,shrikumar2017deeplift,ancona2017better} and perturbation-based methods, such as LIME and anchors \cite{ribeiro2016lime, ribeiro2018anchors}. The main problem with these methods is implementational. Backpropagation-like models can be used only on neural networks (or other differentiable models) with known architectures, and are therefore not suitable to try to interpret  black-box models. Perturbation-based methods, while intuitive at first sight, may actually be cumbersome to use. In fact, they require the user to understand the topology of the input space in order to define an appropriate neighborhood of a sample so as to find meaningful perturbations \cite{molnar2019, fong2017interpretable}.
%
\section{Problem formulation}\label{sec:formulation}
\subsection{The scope of interpretability}
In order to properly design an interpretable model, it is necessary to first define the goals of interpretability. In what follows, we propose that any interpretable model should aim to achieve two main goals:
\begin{goal}\label{goal:transparency}
\emph{Understanding a decision process.}  While opaqueness concerning machine behaviour might not always be a problem, in high-stakes scenarios such as healthcare, model interpretability, \eg providing explanations about the clinical and biological factors that are driving the predictions, is crucial to gain the trust of users and third parties affected by the prediction, \eg clinicians and patients. 
\end{goal}
\begin{goal}\label{goal:bias}
\emph{Bias identification.} Many real-world datasets contain biases, \eg \citet{wachinger2018databias}. An interpretable model can potentially unveil biases in these datasets (as in the work of \citet{tan2017detecting}), or in a deployed model (in a post-hoc scenario). This is important in matters of ethics, fairness, and safety among others. 
\end{goal}
\subsection{Desiderata for an interpretable model}\label{sec:desiderata}
\paragraph{Simultaneous transparency and simulatability} Based on the drawbacks of previous models for interpretability reported in Section~\ref{sec:related}, we argue that a model, to be interpretable by a human user, should produce a prediction based on an easily understood and \emph{complete} set of rules (\emph{transparency}) involving as few features as possible (so it can be easily \emph{simulated} by the human user). If this is not possible, (still understandable) high-level features  have to be inferred. This is needed for Goal~\ref{goal:transparency}.
\paragraph{Expressiveness} In order to identify biases in a dataset or in other models (Goal~\ref{goal:bias}), the interpretable model should be unbiased, \ie an universal approximator. To understand this, consider, for example, a dataset used for prediction of violent crimes with a strong bias towards a certain ethnicity. If an interpretable model has an a priori (inductive) bias towards objects held in hands, then it would not be able to detect the (unfair) bias characterizing this dataset. On the other hand, if we use an unbiased interpretable model, it could learn the correlation between ethnicity and crimes inherent in the dataset. By inspecting the explanations provided by the interpretable model, we would then be able to identify this dataset bias.
%
%
\subsection{The case for monotonicity-constrained networks}\label{sec:motivation}
In this paper we claim that the  two desiderata in Section~\ref{sec:desiderata} can be achieved by a special class of neural networks, where the layers from the input to a chosen hidden layer (say $k$) are left unconstrained, while the  layers from $k$ to the output are built  to enforce a monotonic relationship between the layer $k$ and the output layer (where monotinicity is defined as in Definition~\ref{def:mono_multi_multi}). This construction allows the layer $k$ to learn arbitrary high-level features. We argue that these features are in a certain sense interpretable.\par

Our argumentation is based on the comparison of  linear classifiers to nonlinear ones. Linear classifiers are generally regarded as interpretable methods because users can trivially understand if an output \emph{increases} or \emph{decreases} when a  predictor  is changed. On the other hand, in nonlinear classifiers the possible correlations among input variables makes it difficult to predict how the output would change if a single variable changes. Enforcing monotonicity allows us to reason about the behavior of the output \wrt a single predictor independently from the others, granting us a certain degree of intuition about the  model predictions. 
\subsection{Notation and definitions}\label{sec:notation}
We represent vectors with lowercase boldface letters, \eg $\mathbf{x}$, and matrices with uppercase boldface letters, \eg $\mathbf{W}$. Elements of vectors and matrices are denoted with lowercase subscripts. The $i$-th element of vector $\mathbf{x}$ is $x_i$ and the element in row $i$ and column $j$ of matrix $\mathbf{W}$ is $W_{ij}$.
Given a function $\mathbf{y}=f(\mathbf{x})$, we denote by $\frac{\partial y_i}{\partial x_j}$ the partial derivative of the $i$-th component of $\mathbf{y}$ \wrt the $j$-th component of $\mathbf{x}$.

In this work we focus on multilayer perceptrons (MLPs). Let the 
function $\mathbf{y}=f(\mathbf{x})$ implement a MLP with $L+1$ layers. We denote a layer $k$ of the network by $\mathbf{h}^{(k)}$ for $k=0, \dots , L$. In particular, the input $\mathbf{x}$ and the output $\mathbf{y}$ correspond, respectively, to $\mathbf{h}^{(0)}$ and $\mathbf{h}^{(L)}$. As is customary, we represent a nonlinearity as $\sigma(\cdot)$. We can therefore write $\mathbf{h}^{(k+1)} = \sigma(\mathbf{W}^{(k)}\mathbf{h}^{(k)} + \mathbf{b}^{(k)})$, where $\mathbf{W}^{(k)}$ is the weight matrix and $\mathbf{b}^{(k)}$ is the bias. \par
Since there are multiple ways of defining an order in $\mathbb{R}^n$, we shall clarify which notion of monotonicity we are working with.

\begin{definition}\label{def:mono_multi}
A function $f: \mathbb{R}^n \rightarrow \mathbb{R}$ is called \textbf{monotonically increasing} (or non-decreasing) if for all $i=1,\dots,n$ the (univariate) restriction $f\mid_{i}: x_i  \mapsto y = f(\tilde{x}_1,\dots,x_i,\dots,\tilde{x}_n)$, obtained by fixing all the components except  the $i$-th, is monotonically increasing for every fixed value $\tilde{x}_j$ $\forall j = 1,\dots,n$ and $j\neq i$, in the usual sense of monotonicity for univariate functions. The definition for monotonically decreasing functions is analogous.
\end{definition}


\begin{definition}\label{def:mono_multi_multi}
A multivalued function $f: \mathbb{R}^n \rightarrow \mathbb{R}^m$ is called \textbf{monotonically increasing} if every component $f_i$ with $i = 1,\dots,m$ is monotonic according to Definition~\ref{def:mono_multi}.
\end{definition}

\section{Monotonic Features}\label{sec:mono_feat}
Suppose that we want to interpret the prediction of an MLP with $L+1$ layers \wrt its $k$-th layer. In this section we present a way to constrain the $N^{(k)}$ units of the chosen interpretable layer $\mathbf{h}^{(k)} \in \mathbb{R}^{N^{(k)}}$ to be monotonic \wrt the units of the output layer $\mathbf{y} \in \mathbb{R}^{N^{(L)}}$. As mentioned before, in this work we focus only on MLPs. However, our strategy can be easily extended to different architectures. We refer to our neural network model with monotonicity constraints as \emph{MonoNet}.

\subsection{Monotonically increasing layers}\label{sec:exp_layers}
The first step in our construction is building monotonically increasing  layers. We follow the same idea as in \cite{daniels2010monotone, sill1998monotonic}. As discussed in \ref{sec:notation}, the $\mathbf{h}^{(k+1)}$ layer can be computed from the elements of layer  $\mathbf{h}^{(k)}$  as  $\mathbf{h}^{(k+1)} = \sigma(\mathbf{W}^{(k)}\mathbf{h}^{(k)} + \mathbf{b}^{(k)})$. We can now compute the partial derivatives of this relationship as:
\begin{equation}\label{eq:partial_der}
\frac{\partial h^{(k+1)}_i}{\partial h^{(k)}_j} = \underbrace{\sigma'\Big(\sum_{t=1}^{N^{(k)}}W^{(k)}_{it}h^{(k)}_t + b^{(k)}_i\Big)}_{\geq 0}W^{(k)}_{ij}.
\end{equation}
The most commonly used nonlinearities are non-decreasing functions, whose derivatives are always non-negative. Hence, the partial derivative in (\ref{eq:partial_der}) will be non-negative if and only if $W^{(k)}_{ij} \geq 0$. That is, $\mathbf{h}^{(k+1)}$ will be monotonically non-decreasing \wrt $\mathbf{h}^{(k)}$ if and only if the weight matrix has only non-negative entries. A way to impose this constraint is to apply to the weights a function with range in the positive numbers, such as the exponential function:
\begin{equation}\label{eq:exp_layers}
    \mathbf{h}^{(k+1)} = \sigma \Big(\mathrm{exp}(\mathbf{W}^{(k)})\mathbf{h}^{(k)} + \mathbf{b}^{(k)}\Big).
\end{equation}
Since the compositions of monotonically non-decreasing functions are also monotonically non-decreasing, we are guaranteed that, by stacking such layers, the last layer $\mathbf{y} = \mathbf{h}^{(L)}$ is monotonically non-decreasing \wrt the interpretable layer $\mathbf{h}^{(k)}$.\par

\subsection{Allowing arbitrary monotonicity}\label{sec:any_mono}

The construction in the previous section enabled us to ensure a monotonically non-decreasing relationship between a chosen interpretable layer and the output. However, we would like \emph{each} component of the interpretable layer $\mathbf{h}^{(k)}$ to have an \emph{arbitrary} monotonic behavior (\ie either increasing or decreasing) \wrt \emph{each} component of the output $\mathbf{y}$. This can be achieved by component-wise rescaling of both $\mathbf{h}^{(k)}$ and $\mathbf{y}$. To see this let us introduce the auxiliary layers $\mathbf{\tilde{h}}^{(k)}$ and $\mathbf{\tilde{y}}$ so that
\begin{equation}
    \mathbf{\tilde{h}}^{(k)} = \boldsymbol{\alpha}\odot\mathbf{h}^{(k)}, \;\; \mathbf{y} = \boldsymbol{\beta}\odot\mathbf{\tilde{y}},
\end{equation}

where $\boldsymbol{\alpha} \in \mathbb{R}^{N^{(k)}}$, $\boldsymbol{\beta} \in \mathbb{R}^{N^{(L)}}$, and $\odot$ denotes a component-wise multiplication. If we stack monotonic layers from $\mathbf{\tilde{h}}^{(k)}$ to $\mathbf{\tilde{y}}$ as explained in Section~\ref{sec:exp_layers}, we find that the partial derivatives are
\begin{equation}\label{eq:any_mono_der}
    \frac{\partial h_i}{\partial y_j} = \underbrace{\frac{\partial \tilde{h}_i}{\partial \tilde{y}_j}}_{\geq 0} \frac{1}{\alpha_i \beta_j},
\end{equation}
 where the partial derivatives in terms of the auxiliary layers are positive. \par

\def\layersep{1.1cm}
\def\nrin{2}
\def\nrhi{3}
\def\nro{2}

\begin{figure}
\centering
\begin{tikzpicture}[shorten >=1pt,->,node distance=\layersep]
     
    \tikzstyle{every pin edge}=[<-,shorten <=1pt]
    \tikzstyle{neuron}=[circle,minimum size=15pt,inner sep=0pt,draw=black]
    \tikzstyle{input neuron}=[neuron,fill=blue!25];
    \tikzstyle{interpretable neuron}=[neuron,fill=blue!25];
    \tikzstyle{explayer neuron}=[neuron,fill=red!25];
    \tikzstyle{output neuron}=[neuron,fill=green!50];
    \tikzstyle{pointwise weights}=[draw=blue!100,fill=blue!100];
    \tikzstyle{positive weights}=[draw=red,fill=red,opacity=1.0];
    \tikzstyle{inactive weights}=[draw=black,fill=black,opacity=0.5];


    \foreach \name / \y in {1,...,3}
        \path[yshift=0.5cm]
            node[neuron] (H-\name) at (\layersep,-\y) {};

    \foreach \name / \y in {1,...,3}
        \path[yshift=0.5cm]
            node[neuron] (J-\name) at (2*\layersep,-\y) {};
            
    \foreach \name / \y in {1,...,3}
        \path[yshift=0.5cm]
            node[interpretable neuron] (K-\name) at (3*\layersep,-\y) {};
            
    \foreach \name / \y in {1,...,3}
        \path[yshift=0.5cm]
            node[explayer neuron] (T-\name) at (4*\layersep,-\y) {};
            
    \foreach \name / \y in {1,...,3}
        \path[yshift=0.5cm]
            node[explayer neuron] (S-\name) at (5*\layersep,-\y) {};
            
    \foreach \name / \y in {1,...,3}
        \path[yshift=0.5cm]
            node[output neuron] (L-\name) at (6*\layersep,-\y) {};


    \foreach \source in {1,...,3}
        \foreach \dest in {1,...,3}
            \path[inactive weights] (H-\source) edge (J-\dest);
            
    \foreach \source in {1,...,3}
        \foreach \dest in {1,...,3}
            \path[inactive weights] (J-\source) edge (K-\dest);
            
    \foreach \source in {1,...,3}
            \path[pointwise weights] (K-\source) edge (T-\source);
        
    \foreach \source in {1,...,3}
        \foreach \dest in {1,...,3}
            \path[positive weights] (T-\source) edge (S-\dest);

    \foreach \source in {1,...,3}
            \path[pointwise weights] (S-\source) edge (L-\source);

\end{tikzpicture}
\caption{Visualization of the monotonicity constraint. The interpretable layer (blue) is free to learn any representation of the previous layers (white). The red layers are guaranteed (Section~\ref{sec:exp_layers}) to be monotonically increasing \wrt each other because of the positive weights (red arrows). Thanks to component-wise rescaling (blue arrows, Section~\ref{sec:any_mono}), the output (green) can have an \emph{arbitrary} monotonic relationship \wrt to the interpretable layer.}\label{fig:mono_constraint}
\end{figure}
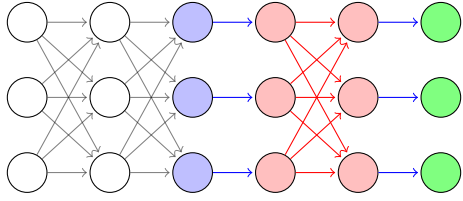

Figure~\ref{fig:mono_constraint} graphically summarizes the construction of a MonoNet.

\subsection{On the representational power of monotone networks}
\citet{daniels2010monotone} proved an analogue of the universal approximation theorem \cite{hornik1991approximation} for monotonically non-decreasing functions. It has to be noted that the definition of monotonicity used by \citet{daniels2010monotone} is more relaxed than ours. Therefore their theorem is valid in particular for our definition of monotonicity (Definition~\ref{def:mono_multi}). Since any monotonically non-increasing function can be obtained by changing the sign of a monotonically non-decreasing function, the result can be extended to monotonically non-increasing functions. 

Now the question is, given a MonoNet with $L+1$ layers, do we still retain the same universal approximation capabilities of neural networks by constraining the output layer $\mathbf{y}$ to be monotonic \wrt $\mathbf{h}^{(k)}$? The answer is yes.
This can be understood with a somewhat extreme example. The last layer of deep learning architectures is always monotonic (either linear or with known nonlinearity given by the activation function) according to Definition~\ref{def:mono_multi_multi}, and hence can be potentially approximated by our monotonic construction. This means that the apparently-constrained MonoNet can in fact approximate any function that a classic (unconstrained) neural network that has the \emph{same} first $k-1$ layers can approximate. However, this is not the use-case of interest. Instead, our model becomes useful when an inference problem can be solved by learning a hidden representation that has an \emph{arbitrary nonlinear monotonic} relationship \wrt the output.

\subsection{Towards interpretability}\label{sec:towards}
While we argued that monotonicity can improve our understanding of a model, we are still not able to fully interpret a model just by learning monotonic \emph{features}. Ideally, we would like to understand the behavior of the MonoNet \wrt the original \emph{input}. In Section~\ref{sec:experiments} we show that, by computing simple statistics, it is still possible to get an approximate idea of how the monotonic features relate to the input space. 

\subsection{On hierarchical monotonic features}\label{sec:hierarchical}

In the previous sections we presented how to impose a monotonicity constraint between the output layer and a chosen hidden layer. It can be noticed though that it is possible to stack several monotone (sub-)networks to form a hierarchy of monotonic features. Note, however, that monotonicity is satisfied only for layers directly connected by a single monotone sub-network. Nonetheless, such hierarchy could help us to think about the learned features in a modular way, similarly to how we would inspect a decision tree level-by-level. The hope is to learn a hierarchy of increasingly complex representations like in classic neural networks \cite{zeiler2014visualizing} with the advantage of better interpretability given by the monotonicity constraint. Furthermore, with this construction we may be able to get to a closer interpretation \wrt the input space (Section~\ref{sec:towards}).

\section{Experimental validation}\label{sec:experiments}


\subsection{Understanding the interpretable features}\label{sec:understanding_features}
As mentioned in Section~\ref{sec:towards}, MonoNets are not directly interpretable \wrt to the input space. It is however possible to get an approximate idea of the patterns in the input space that most (or least) activate a unit  of the interpretable layer $\mathbf{h}^{(k)}$. To see this, let us consider, for example, the unit $h^{(k)}_i$ of $\mathbf{h}^{(k)}$ and assume that a certain unit $y_j$ of the output layer $\mathbf{y}$ (which could represent, for instance, the probability of one class or a regressed value) is increasing \wrt $h^{(k)}_i$. This means that if we order the training samples according to the values of $h^{(k)}_i$ we can potentially ``unveil'' a feature that positively correlates with $y_j$. In this work, we try to unveil these features by analysing the top and bottom distributions of the samples ranked according to the interpretable features. In Section~\ref{sec:interpret_comparison}, we present  a working example of this concept.

\subsection{Interpreting Risk Score Prediction}\label{sec:risk_pred}

We compare our model against models that are regarded as interpretable: {\fontfamily{cmtt}\selectfont risk-slim} \cite{ustun2016learning, ustun2017kdd} and decision trees \cite{quinlan1986decision}. 
Given a dataset, {\fontfamily{cmtt}\selectfont risk-slim} computes a score for each predictor (\eg Table~\ref{table:slim_income}). At inference time, for each sample a total score $S$ is computed by summing the scores of the features characterising the sample. The probability for $y=1$ (which denotes the ``risk'') is then computed as:
\begin{equation}\label{eq:risk_model}
    \mathbb{P}(y=1)=\frac{1}{1+\mathrm{e}^{-(\mathrm{offset} + S)}},
\end{equation}
where the \emph{offset} is learned in conjunction with the feature scores.
We benchmark the models on risk score prediction datasets provided by \citet{ustun2016learning}.\footnote{\href{https://github.com/ustunb/risk-slim/tree/master/examples/data}{https://github.com/ustunb/risk-slim/tree/master/examples/data}}
Table~\ref{table:risk_acc} shows that our model performs similarly to the other models on these risk scores prediction datasets.\par
%

\begin{table}[!htb]\caption{Test accuracies (in \%) of our model, {\fontfamily{cmtt}\selectfont risk-slim}, and decision trees on risk score datasets. Our model exploits a network with layers of size (input-64-64-3-64-1). We choose to interpret the layer of size 3. In the table, we report its average accuracy, with standard deviation, over 10 runs. Best results are reported in bold characters.\newline}\label{table:risk_acc}
\centering
\begin{tabular}{c c c c c c}
\hline
\multirow{2}{*}[-0.3em]{\textbf{Model}} & \multicolumn{5}{c}{\textbf{Dataset}} \\
\cmidrule(lr){2-6}
         & \emph{income} &  \emph{mammo} &  \emph{mushroom} & \emph{breast} & \emph{bank} \\ \hline
{\fontfamily{cmtt}\selectfont risk-slim}  & 75.31 & 53.61 & \textbf{100.00} & 84.06 & 61.30 \\ 
Decision Tree  & 82.16 & \textbf{76.29} & 96.92 & 94.20 & 57.49 \\ 
MonoNet   & \textbf{84.29} $\pm$ 0.16 & 71.65 $\pm$ 7.67 & 96.01 $\pm$ 0.65 & \textbf{95.79} $\pm$ 2.19 & \textbf{63.05} $\pm$ 1.41\\ \hline
\end{tabular}
\end{table}
\subsubsection{Interpretable features: comparison against {\fontfamily{cmtt}\selectfont risk-slim}} \label{sec:interpret_comparison}
We report the decision rules learned by the risk score models and our model in Table~\ref{table:risk_results} on the \emph{income} and the \emph{mushroom} datasets. For our model, we build bottom and top distributions for each interpretable feature, as explained in Section~\ref{sec:understanding_features} and report the 4 predictors with the biggest gap between the distributions' means. \par
In the \emph{income} dataset, each sample is an adult with  demographics information, such as gender, working hours, education. The task is to predict whether the person is earning more than 50K dollars. In the \emph{mushroom} dataset, the task is to predict if a mushroom is poisonous using some of its features, \eg smell, shape, colour. In both datasets, predictors and outcomes are binary.\footnote{Please refer to the website provided for further information about the datasets, such as feature names.} \par
From the name of the features in the \emph{income} dataset, it is reasonable to believe that many predictors are strongly correlated, \eg married vs. not married .  Indeed, a correlation analysis confirms this hypothesis. 
Interestingly, MonoNet was able to learn a feature (top row in Table~\ref{table:our_income}) that consistently ranks the samples according to gender and marital status (top distribution: never married females, bottom distribution: married males). That is, when ranking the training samples according to the value they assume in the first unit $h^{(k)}_1$ of the interpretable layer, most of the samples with high $h^{(k)}_1$ value are females that never married. Conversely, most of the samples with low $h^{(k)}_1$ are married males. According to our model, this feature is negatively correlated with the outcome, implicating that MonoNet gives married men a higher probability of earning more than 50K dollars. This is consistent with the results provided by {\fontfamily{cmtt}\selectfont risk-slim}. In fact, MonoNet seems to uncover features similar to the decision rules learned by {\fontfamily{cmtt}\selectfont risk-slim} (Table~\ref{table:our_income} vs. Table~\ref{table:slim_income}). 

For the \emph{mushroom} dataset, both models agree that the odor (foul vs. none) of the mushroom is an important feature. A deeper analysis reveals that \coolfont{population\_eq\_several} is correlated with \coolfont{gill\_size\_eq\_broad} ($\rho = -0.5064$) according to the Spearman rank correlation coefficient \cite{spearman}. However, the two models seem to disagree on the importance of other predictors. The apparent disagreement might be simply explained by the fact that MonoNet performs worse (Table~\ref{table:risk_acc}) than {\fontfamily{cmtt}\selectfont risk-slim} on the mushroom dataset, and this might actually be because of poorly learned decision rules. 
%

\begin{table}[!htb]
    \caption{Comparison of rules learned by {\fontfamily{cmtt}\selectfont risk-slim} \cite{ustun2016learning, ustun2017kdd} and MonoNet. For MonoNet, the colored cell indicates if the learned feature is positively (green) or negatively (red) correlated with the outcome $y=1$. In the \emph{income} dataset, $y=1$ means a person is earning more than 50K dollars. In the \emph{mushroom} dataset, $y=1$ means the mushroom is poisonous.}\label{table:risk_results}
    \centering
    \begin{subtable}{.4\linewidth}
      \centering
        \caption{{\fontfamily{cmtt}\selectfont risk-slim} on the \emph{income} dataset.}\label{table:slim_income}
        \begin{tabular}{ c  c }
        \hline
            \textbf{Feature} & \textbf{Points}\\\hline
            \coolfont{Age$\leq$21} & $+3$\\
            \coolfont{Married} & $+2$\\
            \coolfont{AnyCapitalGains} & $+2$ \\
            \coolfont{JobManagerial} & $+1$\\
            \coolfont{HSDiploma} & $-1$\\
            \coolfont{NoHS} & $-2$\\
        \end{tabular}
    \end{subtable} 
    \begin{subtable}{.5\linewidth}
      \centering
        \caption{{\fontfamily{cmtt}\selectfont risk-slim} on the \emph{mushroom} dataset.}\label{table:slim_mushroom}
        \begin{tabular}{ c  c }
        \hline
            \textbf{Feature} & \textbf{Points}\\\hline
            \coolfont{odor\_eq\_foul} & $+5$\\
            \coolfont{spore\_print\_color\_eq\_green} & $+5$\\
            \coolfont{gill\_size\_eq\_broad} & $-4$\\
            \coolfont{odor\_eq\_almond} & $-5$\\
            \coolfont{odor\_eq\_anise} & $-5$\\
            \coolfont{odor\_eq\_none} & $-5$\\
        \end{tabular}
    \end{subtable} 
    
    \centering
    \begin{subtable}{\linewidth}\centering
        \caption{MonoNet on the \emph{income} dataset.}\label{table:our_income}
        \begin{tabular}{ c c }
            \hline
            \textbf{Top} & \textbf{Bottom}  \\
            \hline\rowcolor{red!50}
            \coolfont{Female, NeverMarried}  & \coolfont{Married, Male}\\
            \rowcolor{green!50}
            \coolfont{Married} & \coolfont{HSDiploma, NeverMarried, JobService}\\
            \rowcolor{green!50}
            \coolfont{Married} & \coolfont{NeverMarried, WorkHrsPerWeek$\leq$40, 22$\leq$Age$\leq$29}\\
            \hline
        \end{tabular}
    \end{subtable}
    
    \centering
    \begin{subtable}{\linewidth}\centering
        \caption{MonoNet on the \emph{mushroom} dataset.}\label{table:our_mushroom}
        \begin{tabular}{ P{0.42\linewidth} P{0.42\linewidth}}
            \hline
            \textbf{Top} & \textbf{Bottom}\\
            \hline\rowcolor{red!50}
              \coolfont{stlk\_color\_above\_ring\_eq\_white} & \coolfont{stlk\_root\_eq\_bulbous, population\_eq\_several, stlk\_color\_below\_ring\_eq\_pink}\\
            \rowcolor{red!50}\coolfont{odor\_eq\_none} & \coolfont{odor\_eq\_foul, stlk\_srfce\_blw\_ring\_eq\_grooves, stlk\_srfce\_abv\_ring\_eq\_grooves}\\
            \rowcolor{green!50}\coolfont{odor\_eq\_foul, stlk\_srfce\_abv\_ring\_eq\_grooves} & \coolfont{odor\_eq\_none, ring\_type\_eq\_pendant}\\
            \hline
        \end{tabular}
    \end{subtable}
\end{table}

\subsection{Understanding a model with hierarchical monotone features}\label{sec:image_class}

\begin{figure}[!htb]
    \centering
    
    \begin{tikzpicture}[x={(1,0)},y={(0,1)},z={({cos(60)},{sin(60)})},
font=\sffamily\small,scale=2]
%
\draw pic (box1) at (0.0,-0.1,0) {fake box=white!70!gray with dimensions 0.4 and {1.9} and 0.8};

\foreach \X/\Col in {0.9/red,1.1/white,1.3/green,1.4/blue}
{\draw[canvas is yz plane at x = \X, transform shape, draw = black, fill =
\Col!50!white, opacity = 0.5] (0.6,-0.2) rectangle (2.1,-1.5);}
%
\node[draw,single arrow, black,fill=white!100] at (1.4,0.6,0) {MAX};
\node[circle,draw,blue,fill=blue!30] (A1) at (2.0,1,0) {~~~};
\node[circle,draw,green,fill=green!30,below=4pt of A1] (A2) {~~~};
\node[circle,draw,red,fill=red!30,below=18pt of A2] (A3) {~~~};
\draw[circle dotted, line width=2pt,shorten <=3pt] (A2) -- (A3);
\node[circle,draw,gray] (B1) at (2.4,1,0) {~~~};
\node[circle,draw,gray,below=4pt of B1] (B2) {~~~};
\node[circle,draw,gray,below=18pt of B2] (B3) {~~~};
\draw[circle dotted, line width=2pt,shorten <=3pt] (B2) -- (B3);
\node[circle,draw,gray,fill=white!100] (C1) at (2.8,1,0) {~~~};
\node[circle,draw,gray,fill=white!100,below=4pt of C1] (C2) {~~~};
\node[circle,draw,gray,fill=white!100,below=18pt of C2] (C3) {~~~};
\draw[circle dotted, line width=2pt,shorten <=3pt] (C2) -- (C3);
\node[circle,draw,gray] (D1) at (3.2,1,0) {~~~};
\node[circle,draw,gray,below=4pt of D1] (D2) {~~~};
\node[circle,draw,gray,below=18pt of D2] (D3) {~~~};
\draw[circle dotted, line width=2pt,shorten <=3pt] (D2) -- (D3);
\node[circle,draw,gray,fill=white!] (E1) at (3.6,1,0) {~~~};
\node[circle,draw,gray,fill=white!,below=4pt of E1] (E2) {~~~};
\node[circle,draw,gray,fill=white!,below=18pt of E2] (E3) {~~~};
\draw[circle dotted, line width=2pt,shorten <=3pt] (E2) -- (E3);
\begin{scope}[on background layer]
\node[gray,thick,rounded corners,fill=gray!50,fit=(A1) (A3)]{};
\node[gray,thick,rounded corners,fill=gray!50,fit=(C1) (C3)]{};
\node[gray,thick,rounded corners,fill=gray!50,fit=(E1) (E3)]{};
\end{scope}
\foreach \X in {1,2,3}
{\draw[-latex] (A\X) -- (B\X);}
\end{tikzpicture}
\caption{Architecture to learn hierarchical monotonic features. Each activation maps (colored planes) generated by each convolutional filter are summarized in a single unit (colored circles). These ``summaries'' are monotonic \wrt a hidden layer, which in turn is monotonic \wrt the output. The layers on which the monotonicity constraint is enforced are denoted by the gray background.}\label{fig:interpret_convnet}
\end{figure}
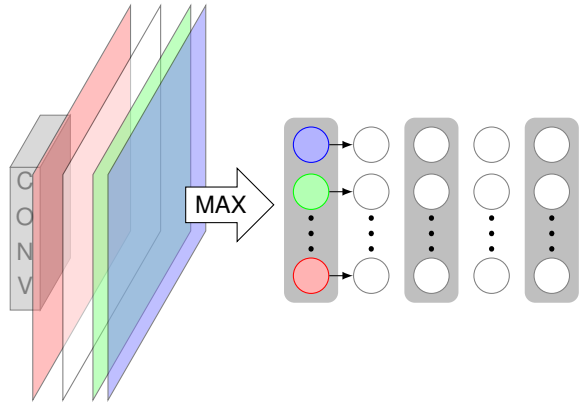

Here we illustrate how a model could be \emph{interpreted} using hierarchical monotonic features (Section~\ref{sec:hierarchical}) with a study on the MNIST dataset. The idea is to enforce the monotonicity constraint between the convolutional filters and a hidden interpretable layer, which is monotonic \wrt to the output. To facilitate the interpretation of the hidden features \wrt the filters, we summarize the activation maps generated by each filter in a single number per filter using a max-pooling operation. These ``summaries'' will form the first layer of hierarchical interpretable features. The architecture is shown in Figure~\ref{fig:interpret_convnet}. 

\begin{figure}
     \centering
     \begin{subfigure}[t]{0.2\textwidth}
         \centering
         \includegraphics[width=\textwidth]{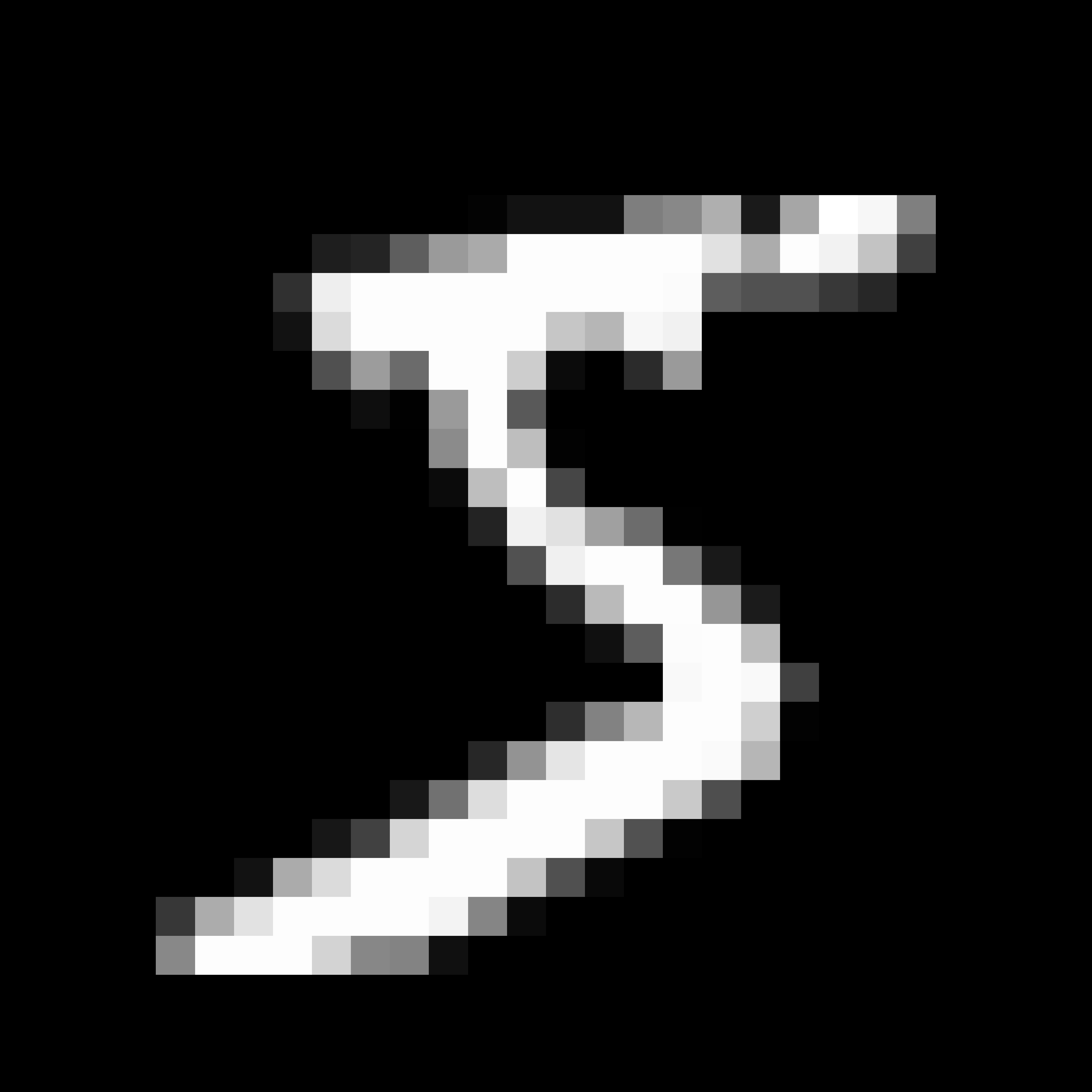}
         \caption{}
         \label{fig:wrong_class}
     \end{subfigure}
     \quad
     \begin{subfigure}[t]{0.2\textwidth}
         \centering
         \includegraphics[width=\textwidth]{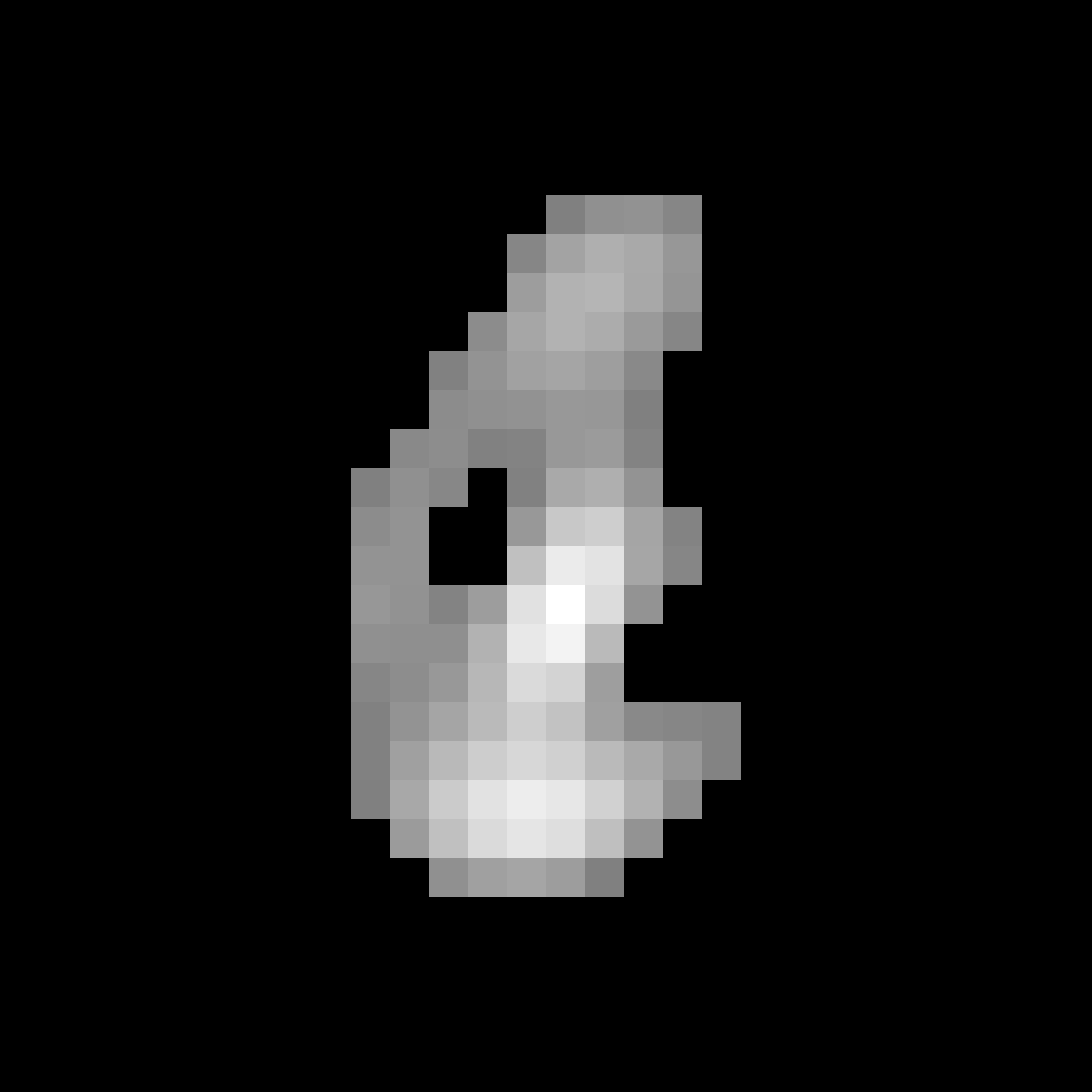}
         \caption{}
         \label{fig:hidden_interpret}
     \end{subfigure}
     \quad
     \begin{subfigure}[t]{0.2\textwidth}
         \centering
         \includegraphics[width=\textwidth]{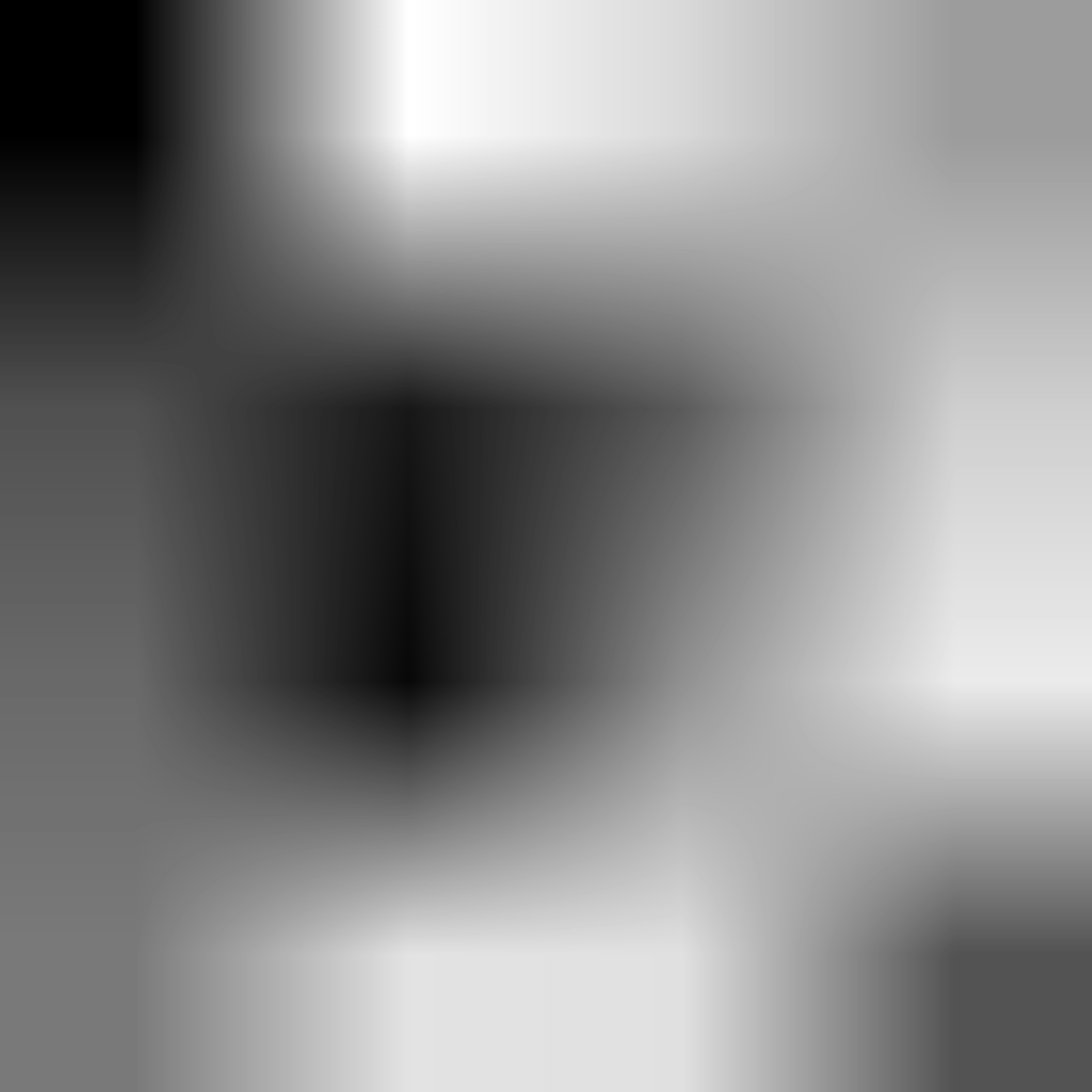}
         \caption{}
         \label{fig:filter_interpret}
     \end{subfigure}
     \quad
     \begin{subfigure}[t]{0.2\textwidth}
         \centering
         \includegraphics[width=\textwidth]{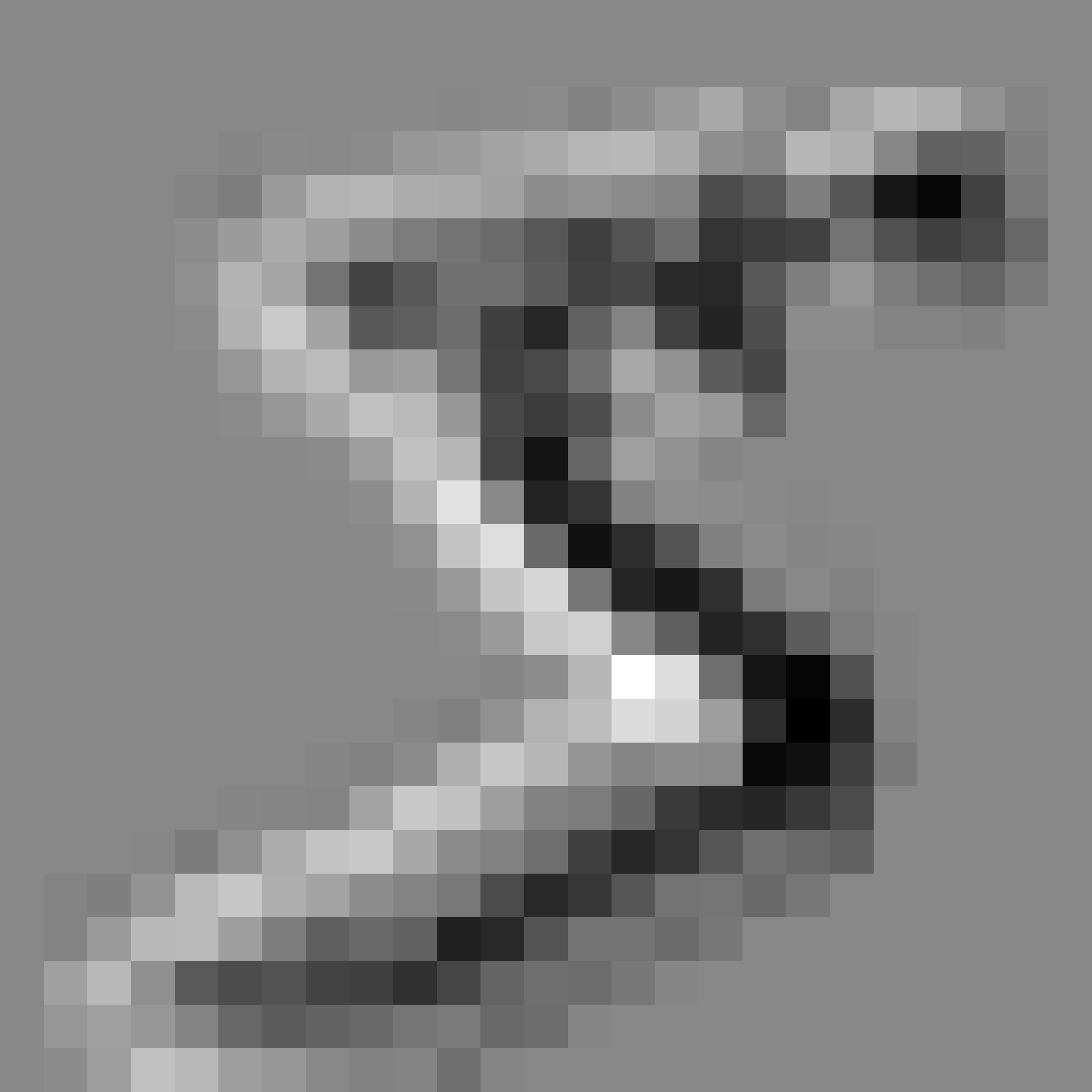}
         \caption{}
         \label{fig:activation_map}
     \end{subfigure}
        \caption{Example of the interpretation process presented in Section~\ref{sec:image_class}. (a) A misclassified sample. True class: 5. Predicted: 3. (b) A high-level monotonic feature learned in a hidden layer. This feature is positively correlated to class 3. (c) A convolution filter which is monotonic \wrt to (b). (d) Activation map generated by (c). White means higher value.}
        \label{fig:understanding_hierarchical}
\end{figure}

Figure~\ref{fig:wrong_class} shows a sample that a trained model (with the architecture presented above) misclassified. To understand why this happened, we can inspect the features that are monotonically increasing \wrt the wrong class (Figure~\ref{fig:hidden_interpret}). Iteratively, to understand why such a feature might have been ``activated'', we focus on those filters that are monotonically increasing \wrt it (\eg Figure~\ref{fig:filter_interpret}). Finally, we can identify which part of the image contributed to the wrong classification by examining the activation map of the filter. \par
We stress the fact that this seemingly trivial reasoning to unveil the decision process of the network was possible because of the monotonicity constraint. This allowed us to reason about particular features, \emph{independently} from the other features (assuming that they are kept \emph{fixed}, see Definition~\ref{def:mono_multi}). In an unconstrained neural network, we would need to know the actual values of the other features to know if a feature of interest is contributing positively or negatively towards a prediction. \par
As a final remark, we note that one might be tempted to modify the original image around the region detected on the activation map of a filter. This, however, would have unpredictable results since 
the input space itself is not monotone \wrt the output.

\section{Discussion}\label{sec:discussion}
\paragraph{Summary} In this work we proposed some desiderata for an interpretable model. Based on these assumptions, we suggested that learning monotonic features in a neural network can lead to models that can be considered interpretable to a certain extent. We demonstrated how monotic features can be obtained by structurally constraining a MLP. 
Our model, MonoNet, shows promising results. However, it is not yet ``fully'' interpretable: while the monotonicity constraint helps us to interpret predictions in terms of the learned hidden features, we  would ultimately like to interpret the predictions \wrt the input space. We proposed and experimentally validated two approaches towards solving this issue: ranking \wrt monotonic features values and hierarchical monotonic features. 

\paragraph{Comparison with self-explaining neural networks (SENN)} \citet{melis2018towards} recently proposed another neural network architecture with built-in interpretability. Their work is  similar in spirit to ours. They also design an architecture able to learn high level features that are monotonically related to the output. However, this monotonic relation is restricted to being additively separable. In this regard, our work can be considered as an extension of theirs. The advantage of their method, though, is that for each high-level feature, they are able to learn an importance score. This score is learned by imposing a ``local explainability'' constraint during training. This, together with the additive separability condition mentioned above, may however introduce a bias in their model, which we have claimed is not desirable for an interpretable method.

\bibliographystyle{unsrtnat}
\bibliography{references}

\end{document}